# Exhaustive and Efficient Constraint Propagation: A Semi-Supervised Learning Perspective and Its Applications

Zhiwu Lu, Horace H.S. Ip, and Yuxin Peng

*Abstract*—This paper presents a novel pairwise constraint propagation approach by decomposing the challenging constraint propagation problem into a set of independent semi-supervised learning subproblems which can be solved in quadratic time using label propagation based on $k$-nearest neighbor graphs. Considering that this time cost is proportional to the number of all possible pairwise constraints, our approach actually provides an efficient solution for exhaustively propagating pairwise constraints throughout the entire dataset. The resulting exhaustive set of propagated pairwise constraints are further used to adjust the similarity matrix for constrained spectral clustering. Other than the traditional constraint propagation on single-source data, our approach is also extended to more challenging constraint propagation on multi-source data where each pairwise constraint is defined over a pair of data points from different sources. This multi-source constraint propagation has an important application to cross-modal multimedia retrieval. Extensive results have shown the superior performance of our approach.

## I. INTRODUCTION

In computer vision and multimedia content analysis, much effort has been made to handle different challenging problems by developing new machine learning techniques. Although encouraging results have been reported in the literature, these techniques suffer from severe performance degradation in practice, due to complicated data structures, inherent modeling limitations and so on. Therefore, any extra supervisory information must be exploited to reduce such performance degradation and improve the quality of machine learning. The labels of data points are potential sources of such supervisory information which has been widely used. In this paper, we consider a commonly adopted and weaker type of supervisory information, called pairwise constraints which specify whether a pair of data points occur together.

There exist two types of pairwise constraints, known as *must-link* constraints and *cannot-link* constraints, respectively. We can readily derive such pairwise constraints from the labels of data points, where a pair of data points with the same label denotes must-link constraint and cannot-link constraint otherwise. It should be noted, however, that the inverse may

Z. Lu and Y. Peng are with the Institute of Computer Science and Technology, Peking University, Beijing 100871, China (e-mail: luzhiwu@icst.pku.edu.cn, pengyuxin@icst.pku.edu.cn).

H. Ip is with the Department of Computer Science, City University of Hong Kong, Kowloon, Hong Kong (e-mail: cship@cityu.edu.hk).

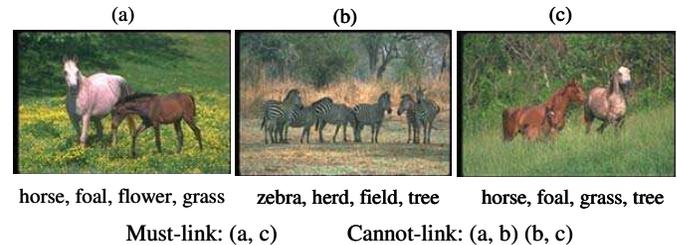

|  (a)  |  (b)  |  (c)  |

horse, foal, flower, grass | zebra, herd, field, tree | horse, foal, grass, tree

Must-link: (a, c)     Cannot-link: (a, b) (b, c)

Fig. 1. The must-link and cannot-link constraints derived from image annotations. Since we focus on recognizing the objects of interests in images, these pairwise constraints are formed without considering the backgrounds such as tree, grass, and field.

not be true, i.e. in general we cannot infer the labels of data points from pairwise constraints, particularly for multi-class datasets. This implies that pairwise constraints are inherently weaker but more general than the labels of data points. Moreover, pairwise constraints can also be automatically derived from domain knowledge [1], [2] or through machine learning. For example, we can obtain pairwise constraints from the annotations of the images shown in Fig. 1. Since we focus on recognizing the objects of interests (e.g. horse and zebra) in images, the pairwise constraints can be formed without considering the backgrounds such as tree, grass, and field. In practice, the objects of interest in images can be roughly distinguished from the backgrounds according to the ranking scores of annotations learnt automatically by an image search engine such as Google.

Pairwise constraints have been widely used for many machine learning problems such as constrained clustering [1]–[4] and metric learning [5]–[7], and it has been reported that the use of appropriate pairwise constraints can often lead to improved results. In this paper, for the convenience of clarifying our motivation, we focus on constrained spectral clustering, i.e., the exploitation of pairwise constraints for spectral clustering [8]–[11] which constructs a new low-dimensional data representation for clustering using the leading eigenvectors of the similarity matrix. Since pairwise constraints specify whether a pair of data points occur together, they provide a source of information about the data relationships, which can be readily used to adjust the similarities between data points for spectral clustering. In fact, constrained spectral clustering has been extensively studied previously. For example, [12] trivially adjusted the similarities between data points to 1 and 0 for must-link and cannot-link constraints, respectively.

This method only adjusts the similarities between constrained data points. In contrast, [13] propagated pairwise constraints to other similarities between unconstrained data points using Gaussian process. However, as noted in [13], this method makes certain assumptions for constraint propagation specially with respect to two-class problems, although a time-consuming heuristic approach for multi-class problems is also discussed. Furthermore, such constraint propagation is also formulated as a semi-definite programming (SDP) problem in [14]. Although the method is not limited to two-class problems, it incurs extremely large computational cost for solving the SDP problem. In [15], the pairwise constraint propagation is also formulated as a constrained optimization problem, but only must-link constraints can be used for optimization.

To overcome these problems with constrained spectral clustering, we propose an exhaustive and efficient constraint propagation approach [16], which is not limited to two-class problems or using only must-link constraints. Specifically, we decompose the challenging constraint propagation problem into a set of independent semi-supervised learning [17]–[19] subproblems. Through formulating these subproblems uniformly as minimizing a regularized energy functional, we can transform the pairwise constraint propagation into solving a continuous-time Lyapunov equation [20] which occurs in many branches of Control Theory such as optimal control and stability analysis [21]. Considering that directly solving the Lyapunov equation scales polynomially to the data size, we further develop an approximate but efficient algorithm based on $k$-nearest neighbor ($k$-NN) graphs using label propagation introduced in [17]. Since the time complexity of this algorithm is quadratic with respect to the data size $N$ and proportional to the total number of all possible pairwise constraints (i.e. $N(N-1)/2$), it can be considered computationally efficient. As compared to the SDP-based constraint propagation [14] with a time complexity of $O(N^4)$, this algorithm is noted to incur much less time cost. Finally, the resulting exhaustive set of propagated pairwise constraints can be used to adjust the similarity matrix for spectral clustering. Although our constraint propagation and similarity adjustment approaches are proposed in the context of constrained spectral clustering, they can be readily applied to many other machine learning problems initially provided with pairwise constraints.

It should be noted that the aforementioned pairwise constraint propagation is limited to single-source data. That is, each pairwise constraint is defined over a pair of data points from the same source. In this paper, we make further attempt to handle more challenging constraint propagation on multi-source data where each pairwise constraint is defined over a pair of data points from different sources. In this case, pairwise constraints still specify whether a pair of data points occur together and the goal of constraint propagation remains the same (i.e. to propagate the initial pairwise constraints throughout the entire dataset). The main difficulty of multi-source constraint propagation lies in how to propagate pairwise constraints across different sources. Fortunately, this challenging problem can be readily decomposed into a series of two-source constraint propagation subproblems. More importantly, such two-source constraint propagation can be formulated as minimizing a regularized energy functional in a semi-supervised learning perspective, similar to constraint propagation on single-source data. Finally, we succeed in developing a similar efficient algorithm for multi-source constraint propagation. When multiple sources refers to text, image, audio and so on, the output of our multi-source constraint propagation actually denotes the correlation between different media sources. That is, our multi-source constraint propagation can be directly used for cross-modal multimedia retrieval which has drawn much attention recently [22]. For cross-modal retrieval, it is not a feasible solution to combine multiple modalities as previous multi-modal multimedia retrieval methods [23]–[25] did.

In summary, our exhaustive and efficient constraint propagation approach can be seen as a very general technique and has the following advantages:

- This is the first attempt to clearly show how pairwise constraints are propagated throughout the entire dataset in a *semi-supervised learning perspective*. Moreover, the pairwise constraint propagation is first shown to equal to *solving a Lyapunov equation*.
- Different from many previous methods, our approach is *not limited* to two-class problems or using only must-link constraints. Moreover, it *allows soft constraints* [26].
- Although developed in the context of constrained spectral clustering, our approach *has the potential to improve* the performance of many other machine learning techniques.
- When extended to more challenging *multi-source constraint propagation*, our approach can achieve promising results in the application of cross-modal retrieval.

Moreover, upon our short conference version [16], this paper presents three additional contributions: more insightful interpretation of pairwise constraint propagation in a semi-supervised learning perspective (see Section II-A), more explanation of our motivation of adjusting the similarity matrix using the propagated pairwise constraints (see Proposition 3), and nontrivial extension to more challenging multi-source constraint propagation (see Section III).

The remainder of this paper is organized as follows. In Section II, we propose an exhaustive and efficient constraint propagation approach. In Section III, our approach is extended to more challenging multi-source constraint propagation. In Section IV, we present the experimental results to evaluate our approach. Finally, Section V gives the conclusions.

## II. Exhaustive and Efficient Constraint Propagation

This section presents our exhaustive and efficient constraint propagation in detail. We first give our solution to the challenging constraint propagation problem in a semi-supervised learning perspective, and then propose an approximate but efficient algorithm. Finally, we apply the proposed constraint propagation algorithm to constrained spectral clustering.

### A. Problem and Solution

Given a dataset $\mathcal{X} = \{x_1, ..., x_N\}$, we denote a set of initial must-link constraints as $\mathcal{M} = \{(x_i, x_j) : l_i = l_j\}$ and a set of initial cannot-link constraints as $\mathcal{C} = \{(x_i, x_j) : l_i \neq l_j\}$, where $l_i$ is the label of data point $x_i$. As we have mentioned, the two sets of initial pairwise constraints can be directly used to adjust the similarities between data points. In previous work [12], only the similarities between the constrained data points are adjusted, and thus the initial pairwise constraints exert very limited effect on the total similarity adjustment. In this paper, we make attempt to spread the effect of pairwise constraints throughout the entire dataset, thereby enabling the initial pairwise constraints to exert a stronger influence on the total similarity adjustment. For the convenience of description, the exhaustive set of propagated pairwise constraints is denoted as $\mathbf{F} \in \mathcal{F}$, where $\mathcal{F} = \{\mathbf{F} = \{f_{ij}\}_{N \times N} : |f_{ij}| \leq 1\}$. It should be noted that $f_{ij} > 0$ means $(x_i, x_j)$ is a must-link constraint while $f_{ij} < 0$ means $(x_i, x_j)$ is a cannot-link constraint, with $|f_{ij}|$ denoting the confidence score of $(x_i, x_j)$ being a must-link (or cannot-link) constraint. In the following, we will develop a novel pairwise constraint propagation method to find the best solution $\mathbf{F}^* \in \mathcal{F}$ based on the two sets of initial pairwise constraints $\mathcal{M}$ and $\mathcal{C}$.

Fig. 2. Illustration of the matrix $\mathbf{Z}$. When we focus on a single data point (e.g. $x_3$ here), the pairwise constraint propagation can be viewed as a two-class semi-supervised learning problem in both vertical and horizontal directions.

A main obstacle of pairwise constraint propagation lies in that the initial cannot-link constraints are not transitive, particularly for multi-class problems. In this paper, however, we succeed in propagating both must-link and cannot-link constraints throughout the entire dataset. Similar to the representation of the exhaustive set of propagated pairwise constraints, we denote the two sets of initial pairwise constraints $\mathcal{M}$ and $\mathcal{C}$ with a single matrix $\mathbf{Z} = \{z_{ij}\}_{N \times N}$:

$$z_{ij} = \begin{cases} +1, & (x_i, x_j) \in \mathcal{M}; \\ -1, & (x_i, x_j) \in \mathcal{C}; \\ 0, & \text{otherwise.} \end{cases} \tag{1}$$

The above definition is inherently suitable for multi-class problems. We have $|z_{ij}| \leq 1$ for soft constraints [26]. This means that $\mathbf{Z} \in \mathcal{F}$. Since we can directly infer the initial pairwise constraints from $\mathbf{Z}$, the initial pairwise constraints have been represented using $\mathbf{Z}$ without loss of information. An example of $\mathbf{Z}$ is illustrated in Fig. 2.

We make further observations on $\mathbf{Z}$ column by column. It can be observed that the $j$-th column $\mathbf{Z}_{\cdot j}$ actually provides the initial configuration of a *two-class semi-supervised learning problem* with respect to $x_j$ (see Fig. 2), where the "positive class" contains the data points that must appear together with $x_j$ and the "negative class" contains the data points that cannot appear together with $x_j$. More concretely, $x_i$ can be initially regarded as coming from the positive (or negative) class if $z_{ij} > 0$ (or $< 0$), but if $x_i$ and $x_j$ are not constrained thus $x_i$ is initially unlabeled. This configuration of a two-class semi-supervised learning problem is also suitable for soft constraints. According to [17], [18], the semi-supervised learning problem with respect to $x_j$ in vertical direction can be formulated as minimizing a regularized energy functional.

Given the dataset $\mathcal{X}$, we define an undirected weighted graph $G = (V, \mathbf{W})$ with its vertex set $V = \mathcal{X}$ and weight matrix $\mathbf{W} = \{w_{ij}\}_{N \times N}$, where $w_{ij}$ is the weight of the edge between $x_i$ and $x_j$. The weight matrix $\mathbf{W}$ is assumed to be nonnegative and symmetric. The normalized graph Laplacian of $G$ is given by

$$\mathbf{L} = \mathbf{I} - \mathbf{D}^{-1/2} \mathbf{W} \mathbf{D}^{-1/2}, \tag{2}$$

where $\mathbf{I}$ is an $N \times N$ identity matrix and $\mathbf{D}$ is an $N \times N$ diagonal matrix with its $i$-th diagonal element being $\sum_j w_{ij}$. Based on this normalized graph Laplacian $\mathbf{L}$, the pairwise constraint propagation with respect to $x_j$ in vertical direction (see Fig. 2) is formulated as

$$\min_{\mathbf{F}_{\cdot j}} \frac{1}{2} \|\mathbf{F}_{\cdot j} - \mathbf{Z}_{\cdot j}\|_2^2 + \frac{\mu}{2} \mathbf{F}_{\cdot j}^T \mathbf{L} \mathbf{F}_{\cdot j}, \tag{3}$$

where $\mu > 0$ is a regularization parameter and $\mathbf{F}_{\cdot j}$ (or $\mathbf{Z}_{\cdot j}$) is the $j$-th column of $\mathbf{F}$ (or $\mathbf{Z}$). The first term of the above equation denotes the fitting error, which penalizes large changes between the propagated pairwise constraints and the initial ones. The second term denotes an energy functional (also known as the smoothness measure) defined based on the graph $G$, which penalizes large changes between nearby data points. More details of the definition of this energy functional can be found in [18].

Since the other columns of $\mathbf{Z}$ can be handled similarly, we can decompose the pairwise constraint propagation problem in vertical direction into $N$ independent label propagation subproblems which can then be solved in parallel. By merging all of these subproblems into a single optimization problem, the vertical constraint propagation is formulated as:

$$\min_{\mathbf{F}} \frac{1}{2} \|\mathbf{F} - \mathbf{Z}\|_F^2 + \frac{\mu}{2} \text{tr}(\mathbf{F}^T \mathbf{L} \mathbf{F}), \tag{4}$$

where $\text{tr}(\cdot)$ stands for the trace of a matrix. That is, similar to [17], [18], we have formulated the vertical constraint propagation as minimizing a regularized energy functional.

However, it is also possible that a column of $\mathbf{Z}$ contains no pairwise constraints (for example, see the fifth column in Fig. 2). That is, the entries of this column may all be zeros, and for such cases, there is no vertical constraint propagation along this column. We deal with this problem through horizontal constraint propagation, which is performed by considering $\mathbf{Z}$ row by row, instead of column-wise. When we focus on a single data point $x_j$, the pairwise constraint propagation with respect to $x_j$ in horizontal direction (see Fig. 2) can also be viewed as a two-class semi-supervised learning problem and then be formulated similar to equation (3):

$$\min_{\mathbf{F}_{j \cdot}} \frac{1}{2} \|\mathbf{F}_{j \cdot} - \mathbf{Z}_{j \cdot}\|_2^2 + \frac{\mu}{2} \mathbf{F}_{j \cdot} \mathbf{L} \mathbf{F}_{j \cdot}^T, \tag{5}$$

where $\mu > 0$ is a regularization parameter and $\mathbf{F}_{j\cdot}$ (or $\mathbf{Z}_{j\cdot}$) is the $j$-th row of $\mathbf{F}$ (or $\mathbf{Z}$). In fact, the above optimization problem is equivalent to

$$\min_{\mathbf{F}_{j\cdot}^T} \frac{1}{2}\|(\mathbf{F}_{j\cdot}^T) - (\mathbf{Z}_{j\cdot}^T)\|_2^2 + \frac{\mu}{2}(\mathbf{F}_{j\cdot}^T)^T\mathbf{L}(\mathbf{F}_{j\cdot}^T), \qquad (6)$$

which takes the same form as equation (3). If all the $N$ rows are considered together, we can obtain the following horizontal constraint propagation:

$$\min_{\mathbf{F}} \frac{1}{2}\|\mathbf{F} - \mathbf{Z}\|_F^2 + \frac{\mu}{2}\mathrm{tr}(\mathbf{F}\mathbf{L}\mathbf{F}^T). \qquad (7)$$

Finally, the vertical and horizontal constraint propagation can be combined by:

$$\min_{\mathbf{F}} \|\mathbf{F} - \mathbf{Z}\|_F^2 + \frac{\mu}{2}\mathrm{tr}(\mathbf{F}^T\mathbf{L}\mathbf{F} + \mathbf{F}\mathbf{L}\mathbf{F}^T). \qquad (8)$$

The distinct advantage of such combination is that we can propagate the initial pairwise constraints to any pair of data points by considering the two directions simultaneously. That is, the constraint propagation may not break down even if the pairwise constraints are missing for certain data points. Let $\mathcal{Q}(\mathbf{F})$ denote the objective function in equation (8). Differentiating $\mathcal{Q}(\mathbf{F})$ with respect to $\mathbf{F}$ and setting it to zero, we have the following equation:

$$\frac{\partial \mathcal{Q}}{\partial \mathbf{F}} = 2(\mathbf{F} - \mathbf{Z}) + \mu\mathbf{L}\mathbf{F} + \mu\mathbf{F}\mathbf{L} = 0, \qquad (9)$$

which can be transformed into a symmetric form

$$(\mathbf{I} + \mu\mathbf{L})\mathbf{F} + \mathbf{F}(\mathbf{I} + \mu\mathbf{L}) = 2\mathbf{Z}. \qquad (10)$$

In fact, the above equation is a standard continuous-time Lyapunov matrix equation [20] and is used in various areas of Control Theory such as optimal control and stability analysis [21]. According to Proposition 1, the Lyapunov matrix equation (10) has a unique and symmetric solution.

*Proposition 1:* The Lyapunov matrix equation (10) has a unique and symmetric solution.

*Proof:* Since the graph Laplacian $\mathbf{L}$ is nonnegative semidefinite and $\mu > 0$, the matrix $\mathbf{I} + \mu\mathbf{L}$ is positive definite and all of its eigenvalues are positive. Hence, for any pair of eigenvalues of $\mathbf{I} + \mu\mathbf{L}$, i.e. $\lambda_i$ and $\lambda_j$, $\lambda_i + \lambda_j \neq 0$. According to [27], the Lyapunov matrix equation (10) has a unique solution. Moreover, since both $\mathbf{L}$ and $\mathbf{Z}$ are symmetric, we can get $((\mathbf{I} + \mu\mathbf{L})\mathbf{F})^T + (\mathbf{F}(\mathbf{I} + \mu\mathbf{L}))^T = 2\mathbf{Z}^T$, i.e. $\mathbf{F}^T(\mathbf{I} + \mu\mathbf{L}) + (\mathbf{I} + \mu\mathbf{L})\mathbf{F}^T = 2\mathbf{Z}$. This means that $\mathbf{F}^T$ is the solution of the Lyapunov matrix equation (10) if $\mathbf{F}$ is. However, we have proven that this equation has a unique solution. Hence, $\mathbf{F}^T = \mathbf{F}$ if $\mathbf{F}$ is the solution. ∎

In summary, by formulating the pairwise constraint propagation as minimizing a regularized energy functional in a semi-supervised learning perspective, we have, for the first time, shown that pairwise constraint propagation actually equals to solving a Lyapunov equation which has been widely used in Control Theory. That is, we have given rise to interesting insight that links pairwise constraint propagation to the Lyapunov equation. However, directly solving the Lyapunov equation has a polynomial time complexity with respect to the data size, although many numerical methods have been developed in the literature. In the following, we will propose an approximate but efficient algorithm, instead of directly solving the Lyapunov equation.

### B. The Proposed Algorithm

Although the pairwise constraint propagation problem (8) can be handled by solving the Lyapunov matrix equation (10), it leads to a large computational cost. To develop an efficient algorithm, we approximately solve the pairwise constraint propagation problem (8) in two optimization steps: (1) $\mathbf{F}_v^* = \arg\min_{\mathbf{F}} \frac{1}{2}\|\mathbf{F} - \mathbf{Z}\|_F^2 + \frac{\mu}{2}\mathrm{tr}(\mathbf{F}^T\mathbf{L}\mathbf{F})$; (2) $\mathbf{F}^* = \arg\min_{\mathbf{F}} \frac{1}{2}\|\mathbf{F} - \mathbf{F}_v^*\|_F^2 + \frac{\mu}{2}\mathrm{tr}(\mathbf{F}\mathbf{L}\mathbf{F}^T)$. That is, we first perform the vertical constraint propagation and then the horizontal constraint propagation. Based on $k$-nearest neighbor ($k$-NN) graphs, both vertical and horizontal constraint propagation can be solved efficiently using the label propagation technique introduced in [17]. Strictly speaking, this optimization strategy may only find a suboptimal solution for our pairwise constraint propagation given by equation (8). Fortunately, our later experimental results have demonstrated that the solution obtained by this optimization strategy is comparable to that of the Lyapunov matrix equation.

As we have mentioned, a candidate set of propagated pairwise constraints can be denoted as $\mathbf{F} \in \mathcal{F}$, where $\mathcal{F} = \{\mathbf{F} = \{f_{ij}\}_{N \times N} : |f_{ij}| \leq 1\}$. Particularly, $\mathbf{Z} \in \mathcal{F}$, where $\mathbf{Z}$ collects the initial pairwise constraints. To find the best solution $\mathbf{F}^* \in \mathcal{F}$ based on $\mathbf{Z}$, we propose the following approximate algorithm using the above optimization strategy:

(1) Construct a $k$-NN graph by defining its weight matrix $\mathbf{W} = \{w_{ij}\}_{N \times N}$ as: $w_{ij} = \frac{a(x_i,x_j)}{\sqrt{a(x_i,x_i)}\sqrt{a(x_j,x_j)}}$ if $x_j$ ($j \neq i$) is among the $k$-nearest neighbors of $x_i$ and $w_{ij} = 0$ otherwise, where $\mathbf{A} = \{a(x_i,x_j)\}_{N \times N}$ is the kernel matrix defined on the dataset $\mathcal{X}$. Set $\mathbf{W} = (\mathbf{W} + \mathbf{W}^T)/2$ to ensure $\mathbf{W}$ is symmetric.

(2) Compute the matrix $\bar{\mathbf{L}} = \mathbf{D}^{-1/2}\mathbf{W}\mathbf{D}^{-1/2}$, where $\mathbf{D}$ is a diagonal matrix with its $i$-th diagonal element being $\sum_j w_{ij}$.

(3) Iterate $\bar{\mathbf{F}}_v(t+1) = \alpha\bar{\mathbf{L}}\mathbf{F}_v(t) + (1-\alpha)\mathbf{Z}$ for the vertical constraint propagation until convergence, where $\mathbf{F}_v(t) \in \mathcal{F}$ and $\alpha$ is a parameter in the range $(0,1)$.

(4) Iterate $\mathbf{F}_h(t+1) = \alpha\mathbf{F}_h(t)\bar{\mathbf{L}} + (1-\alpha)\mathbf{F}_v^*$ for the horizontal constraint propagation until convergence, where $\mathbf{F}_h(t) \in \mathcal{F}$ and $\mathbf{F}_v^*$ is the limit of $\{\mathbf{F}_v(t)\}$.

(5) Output $\mathbf{F}^* = \mathbf{F}_h^*$ as the final representation of the propagated pairwise constraints, where $\mathbf{F}_h^*$ is the limit of $\{\mathbf{F}_h(t)\}$.

Below we give a convergence analysis of the above constraint propagation algorithm. Since the vertical constraint propagation in Step (3) can be regarded as label propagation, its convergence has been shown in [17]. More concretely, similar to [17], we have $\mathbf{F}_v^* = (1-\alpha)(\mathbf{I} - \alpha\bar{\mathbf{L}})^{-1}\mathbf{Z}$ as the limit of $\{\mathbf{F}_v(t)\}$. Meanwhile, we can directly obtain an analytical solution $\mathbf{F}_v^* = (\mathbf{I} + \mu\mathbf{L})^{-1}\mathbf{Z}$ for the vertical constraint propagation $\mathbf{F}_v^* = \arg\min_{\mathbf{F}} \frac{1}{2}\|\mathbf{F} - \mathbf{Z}\|_F^2 + \frac{\mu}{2}\mathrm{tr}(\mathbf{F}^T\mathbf{L}\mathbf{F})$. Since $\mathbf{L} = \mathbf{I} - \bar{\mathbf{L}}$, the analytical solution equals to $\mathbf{F}_v^* = ((\mu+1)\mathbf{I} - \mu\bar{\mathbf{L}})^{-1}\mathbf{Z}$,

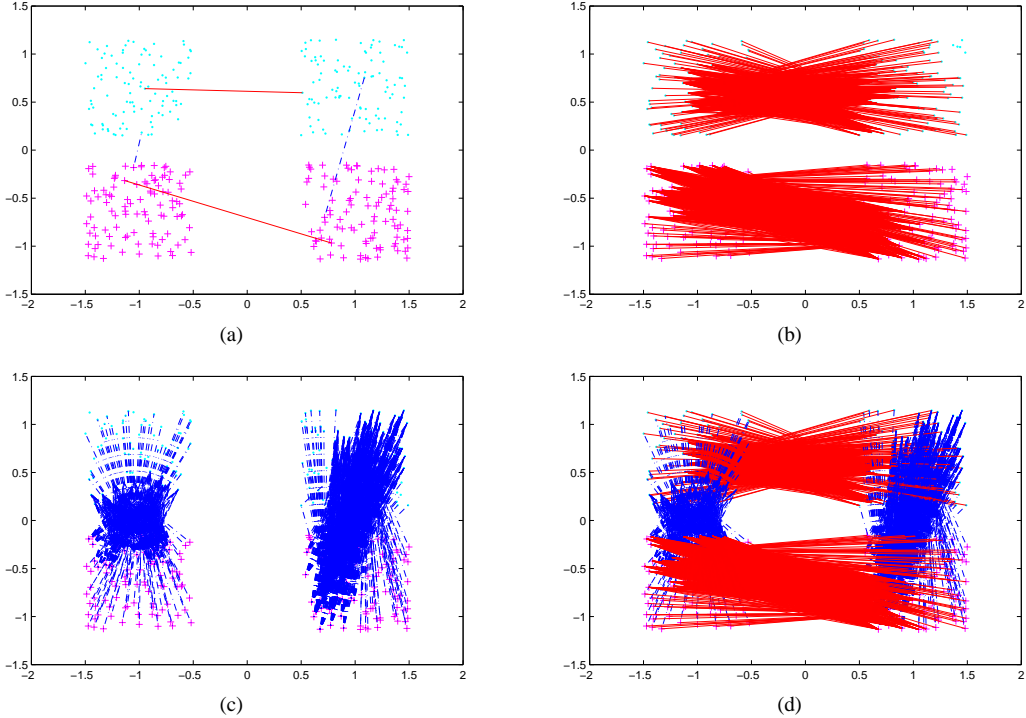

Fig. 3. Illustration of the propagated pairwise constraints obtained by our algorithm: (a) four pairwise constraints and ideal clustering of the dataset; (b) final constraints propagated from only two must-link constraints; (c) final constraints propagated from only two cannot-link constraints; (d) final constraints propagated from four pairwise constraints. Here, must-link constraints are denoted by solid red lines, while cannot-link constraints are denoted by dashed blue lines. Moreover, we only show the propagated constraints with predicted confidence scores > 0.1 in Figs. 3(b)-3(d).

which means that $\alpha = \mu/(\mu + 1)$. As for the horizontal constraint propagation, we have

$$
\begin{aligned}
\mathbf{F}_h^T(t+1) &= \alpha \bar{\mathbf{L}}^T \mathbf{F}_h^T(t) + (1 - \alpha) \mathbf{F}_v^{*T} \\
&= \alpha \bar{\mathbf{L}} \mathbf{F}_h^T(t) + (1 - \alpha) \mathbf{F}_v^{*T}. \quad (11)
\end{aligned}
$$

That is, the horizontal propagation in Step (4) can be transformed into a vertical propagation which converges to $\mathbf{F}_h^{*T} = (1 - \alpha)(\mathbf{I} - \alpha \bar{\mathbf{L}})^{-1} \mathbf{F}_v^{*T}$. Hence, our pairwise constraint propagation algorithm has a closed-form solution as follows:

$$
\begin{aligned}
\mathbf{F}^* &= \mathbf{F}_h^* = (1 - \alpha) \mathbf{F}_v^* (\mathbf{I} - \alpha \bar{\mathbf{L}}^T)^{-1} \\
&= (1 - \alpha)^2 (\mathbf{I} - \alpha \bar{\mathbf{L}})^{-1} \mathbf{Z} (\mathbf{I} - \alpha \bar{\mathbf{L}})^{-1}, \quad (12)
\end{aligned}
$$

which actually accumulates the evidence to reconcile the contradictory propagated constraints for certain pairs of data points. As a toy example, the propagated pairwise constraints given by equation (12) are explicitly shown in Fig. 3. We can find that the propagated pairwise constraints obtained by our algorithm are consistent with the ideal clustering of the dataset.

The above closed-form solution can be further discussed in detail. Firstly, given that both $\bar{\mathbf{L}}$ and $\mathbf{Z}$ are symmetric, this solution is symmetric, just as the solution of the Lyapunov matrix equation (see Proposition 1). Secondly, similar to the above convergence analysis, we can readily obtain the same solution (12), if we first perform the horizontal constraint propagation and then the vertical constraint propagation. To summarize, we have the following proposition:

*Proposition 2:* (i) The closed-form solution (12) is a feasible solution of the Lyapunov matrix equation (10); (ii)

The pairwise constraint propagation in two directions can be alternately performed, no matter which is first.

Although the closed-form solution is only proven to be a feasible solution, our later results have shown that it is comparable to the solution of the Lyapunov matrix equation.

Finally, we give a complexity analysis of our pairwise constraint propagation algorithm. Through semi-supervised learning based on $k$-NN graphs ($k \ll N$), both vertical and horizontal constraint propagation can be performed in quadratic time $O(kN^2)$. Since this time complexity is proportional to the total number of all possible pairwise constraints (i.e. $N(N-1)/2$), our algorithm can be considered computationally efficient. Moreover, our algorithm incurs significantly less computational cost than [14], given that pairwise constraint propagation based on semi-definite programming has a time complexity of $O(N^4)$.

### C. Application to Constrained Spectral Clustering

It should be noted that the output $\mathbf{F}^* = \{f_{ij}^*\}_{N \times N}$ of our constraint propagation algorithm represents an exhaustive set of pairwise constraints with the associated confidence scores $|\mathbf{F}^*|$. Although this exhaustive set of propagated pairwise constraints can be used for many machine learning problems, we focus on constrained spectral clustering whose goal is to obtain a data partition that is fully consistent with the propagated pairwise constraints. More concretely, $\mathbf{F}^*$ can be exploited for constrained spectral clustering by adjusting the original normalized weight matrix $\mathbf{W}$ (i.e. $0 \le w_{ij} \le 1$) of

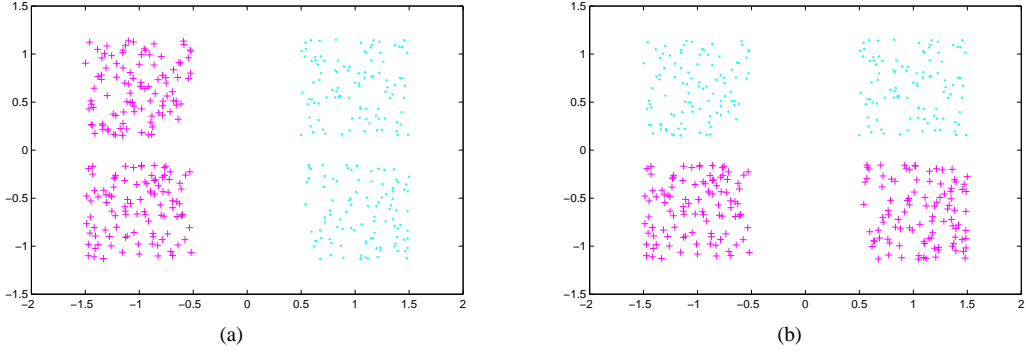

Fig. 4. The results of constrained clustering on the toy dataset using four pairwise constraints given by Fig. 3(a): (a) spectral learning [12]; (b) our approach. The clustering obtained by our approach is consistent with the ideal clustering.

the graph constructed for spectral clustering:

$$\tilde{w}_{ij} = \begin{cases} 1 - (1 - f_{ij}^*)(1 - w_{ij}), & f_{ij}^* \geq 0; \\ (1 + f_{ij}^*)w_{ij}, & f_{ij}^* < 0. \end{cases} \quad (13)$$

The distinct advantage of this weight (or similarity) adjustment strategy is that it can be readily applied to many other machine learning problems provided with pairwise constraints initially.

In the following, the new matrix $\tilde{\mathbf{W}} = \{\tilde{w}_{ij}\}_{N \times N}$ will be directly used for spectral clustering. Here, we need to first prove that $\tilde{\mathbf{W}}$ can be viewed as a new normalized weight matrix by showing that it has the following nice properties.

*Proposition 3:* (i) $\tilde{\mathbf{W}}$ is nonnegative and symmetric; (ii) $\tilde{w}_{ij} \in [0, 1]$; (iii) $\tilde{w}_{ij} \geq w_{ij}$ (or $< w_{ij}$) if $f_{ij}^* \geq 0$ (or $< 0$); (iv) $\frac{\partial \tilde{w}_{ij}}{\partial w_{ij}} = 1 - |f_{ij}^*|$; (v) $\tilde{w}_{ij} \geq f_{ij}^*$ (or $\leq 1 + f_{ij}^*$) if $f_{ij}^* \geq 0$ (or $< 0$).

*Proof:* The above proposition is proven as follows:

(i) The symmetry of both $\mathbf{W}$ and $\mathbf{F}^*$ ensures that $\tilde{\mathbf{W}}$ is symmetric. Since $0 \leq w_{ij} \leq 1$ and $|f_{ij}^*| \leq 1$, we also have: $\tilde{w}_{ij} = 1 - (1 - f_{ij}^*)(1 - w_{ij}) \geq 1 - (1 - w_{ij}) \geq 0$ if $f_{ij}^* \geq 0$ and $\tilde{w}_{ij} = (1 + f_{ij}^*)w_{ij} \geq 0$ if $f_{ij}^* < 0$. That is, we always have $\tilde{w}_{ij} \geq 0$. Hence, $\tilde{\mathbf{W}}$ is nonnegative and symmetric.

(ii) We have proven that $\tilde{w}_{ij} \geq 0$ when $0 \leq w_{ij} \leq 1$ and $|f_{ij}^*| \leq 1$. Similarly, we can prove that $\tilde{w}_{ij} \leq 1$.

(iii) According to equation (13), $\tilde{w}_{ij}$ can be viewed as a *monotonically increasing function* of $f_{ij}^*$. Since $\tilde{w}_{ij} = w_{ij}$ when $f_{ij}^* = 0$, we thus have: $\tilde{w}_{ij} \geq w_{ij}$ (or $< w_{ij}$) if $f_{ij}^* \geq 0$ (or $< 0$).

(iv) $\frac{\partial \tilde{w}_{ij}}{\partial w_{ij}} = 1 - f_{ij}^*$ if $f_{ij}^* \geq 0$, while $\frac{\partial \tilde{w}_{ij}}{\partial w_{ij}} = 1 + f_{ij}^*$ if $f_{ij}^* < 0$. In other words, $\frac{\partial \tilde{w}_{ij}}{\partial w_{ij}} = 1 - |f_{ij}^*|$.

(v) If $f_{ij}^* \geq 0$, $\tilde{w}_{ij}$ can be viewed as a *monotonically increasing function* of $w_{ij}$. Since $\tilde{w}_{ij} = f_{ij}^*$ when $w_{ij} = 0$, we have: $\tilde{w}_{ij} \geq f_{ij}^*$ if $w_{ij} \geq 0$. Moreover, if $f_{ij}^* < 0$, it can be followed from $w_{ij} \leq 1$ and $1 + f_{ij}^* \geq 0$ so that $\tilde{w}_{ij} = (1 + f_{ij}^*)w_{ij} \leq 1 + f_{ij}^*$. ∎

The properties (i) and (ii) show that $\tilde{\mathbf{W}}$ can be used as a normalized weight matrix for spectral clustering. Furthermore, the property (iii) shows that the new weight matrix $\tilde{\mathbf{W}}$ is actually derived from the original weight matrix $\mathbf{W}$ by increasing $w_{ij}$ for the must-link constraints with $f_{ij}^* > 0$ and decreasing $w_{ij}$ for the cannot-link constraints with $f_{ij}^* < 0$. This is consistent with our original motivation of exploiting pairwise constraints for spectral clustering. Although there may exist other weight adjustment methods that also satisfy the property (iii), our approach has two distinct advantages over them given by the properties (iv) and (v), respectively.

More concretely, the property (iv) in Proposition 3 shows the first distinct advantage of our weight adjustment approach, i.e., must-link and cannot-link constraints play the same important role (with the same derivatives) in weight adjustment if they have the same absolute confidence scores, which is reasonable given no prior knowledge. In contrast, although the simple choice by directly setting $\tilde{w}_{ij} = (1 + f_{ij}^*)w_{ij}$ for any $f_{ij}^*$ also has the property (iii), it puts more importance on must-link constraints than on cannot-link constraints (i.e. $1 + |f_{ij}^*| > 1 - |f_{ij}^*|$) even if they have the same absolute confidence scores. As compared with this simple choice, the second advantage of our weight adjustment approach is that $\tilde{w}_{ij}$ can be ensured to be large when $f_{ij}^*$ takes a *large positive value* according to the property (v), even if $w_{ij}$ is initially very small. This is indeed a good thing that only our weight adjustment approach has. Here, it should be noted that both of the two weight adjustment methods can ensure that $\tilde{w}_{ij}$ is small when $f_{ij}^*$ takes a *large negative value* according to the property (v), even if $w_{ij}$ is initially very large.

After we have successfully incorporated the exhaustive set of propagated pairwise constraints obtained by our pairwise constraint propagation algorithm into a new weight matrix $\tilde{\mathbf{W}}$, we then perform spectral clustering with this new weight matrix. The corresponding constrained spectral clustering algorithm is summarized as follows:

(1) Find $K$ largest nontrivial eigenvectors $\mathbf{v}_1, ..., \mathbf{v}_K$ of $\tilde{\mathbf{D}}^{-1/2}\tilde{\mathbf{W}}\tilde{\mathbf{D}}^{-1/2}$, where $\tilde{\mathbf{D}}$ is a diagonal matrix with its $i$-th diagonal element being $\sum_j \tilde{w}_{ij}$.

(2) Form $\mathbf{E} = [\mathbf{v}_1, ..., \mathbf{v}_K]$, and normalize each row of $\mathbf{E}$ to have unit length. Here, the $i$-th row $\mathbf{E}_{i\cdot}$ is the low-dimensional feature vector for data point $x_i$.

(3) Perform $k$-means clustering on the new feature vectors $\{\mathbf{E}_{i\cdot} : i = 1, ..., N\}$ to obtain $K$ clusters.

The clustering results on the toy dataset (see Fig. 3(a)) by the above algorithm are shown in Fig. 4(b). We can find that the clustering obtained by our constraint propagation approach

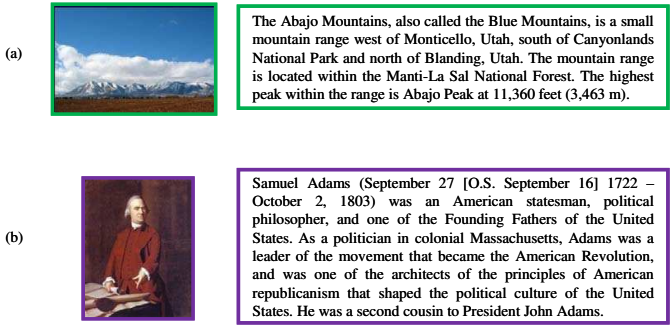

Fig. 5. Illustration of two examples of Wikipedia articles. Each section of a Wikipedia article is associated with a corresponding image. The images and text denote two different data sources.

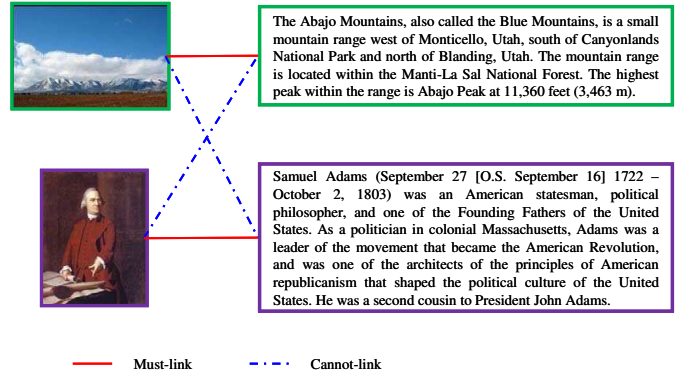

— Must-link    ⋅⋅⋅ Cannot-link

Fig. 6. Illustration of several examples of pairwise constraints defined on a two-source dataset. Both must-link and cannot-link constraints are generated based on the correspondence between images and text shown in Fig. 5.

is consistent with the ideal clustering of the dataset, while this is not true for spectral learning [12] without using constraint propagation (see Fig. 4(a)). In the following, since the pairwise constraints used for constrained spectral clustering (CSC) is obtained by our exhaustive and efficient constraint propagation (E²CP), the above clustering algorithm is also denoted as E²CP to distinguish it from other CSC algorithms.

## III. MULTI-SOURCE CONSTRAINT PROPAGATION

In this section, our E²CP approach is extended to more challenging problem of pairwise constraint propagation on multi-source data. We first give our solution to such multi-source constraint propagation, and then propose an efficient algorithm. Finally, the proposed algorithm is applied to cross-modal multimedia retrieval.

### A. Problem and Solution

We have provided a sound solution to the challenging problem of pairwise constraint propagation in the last section. However, this solution is limited to single-source data. That is, each pairwise constraint is defined over a pair of data points from the same source. In this paper, we further consider even more challenging constraint propagation on multi-source data where each pairwise constraint is defined over a pair of data points from different sources. In this case, pairwise constraints still specify whether a pair of data points occur together and the goal of constraint propagation remains the same (i.e. to propagate the initial pairwise constraints throughout the entire dataset). The main difficulty of multi-source constraint propagation lies in how to propagate pairwise constraints across different data sources. Since this challenging problem can be readily decomposed into a series of two-source constraint propagation subproblems, we focus on two-source constraint propagation and formulate it in detail as follows.

Let $\{\mathcal{X}, \mathcal{Y}\}$ be a two-source dataset, where $\mathcal{X} = \{x_1, ..., x_N\}$ and $\mathcal{Y} = \{y_1, ..., y_M\}$. It should be noted that $\mathcal{X}$ may have a different data size from $\mathcal{Y}$ (i.e. $N \neq M$). As an example, a two-source dataset can be generated with the Wikipedia articles (see Fig. 5), with images and text being the two data sources. For the two-source dataset $\{\mathcal{X}, \mathcal{Y}\}$, we can define a set of initial must-link constraints as $\mathcal{M} = \{(x_i, y_j) : l(x_i) = l(y_j)\}$ and a set of initial cannot-link constraints as $\mathcal{C} = \{(x_i, y_j) : l(x_i) \neq l(y_j)\}$, where $l(x_i)$ (or $l(y_j)$) is the label of $x_i \in \mathcal{X}$ (or $y_j \in \mathcal{Y}$). Here, $x_i$ and $y_j$ are assumed to share the same label set. If the class labels are not provided, the pairwise constraints can be defined only based on the correspondence between two data sources, which can be readily obtained from Web-based content such as Wikipedia articles. Several examples of pairwise constraints defined in this way are illustrated in Fig. 6.

As we have mentioned, the goal of two-source constraint propagation is to propagate the two sets of initial pairwise constraints $\mathcal{M}$ and $\mathcal{C}$ across the two sources $\mathcal{X}$ and $\mathcal{Y}$. In fact, this equals to finding the best solution $\mathbf{F}^* \in \mathcal{F}$ based on $\mathcal{M}$ and $\mathcal{C}$ with $\mathcal{F} = \{\mathbf{F} = \{f_{ij}\}_{N \times M} : |f_{ij}| \leq 1\}$. Here, it should be noted that any exhaustive set of pairwise constraints can be denoted as $\mathbf{F} \in \mathcal{F}$, where $f_{ij} > 0$ means $(x_i, y_j)$ is a must-link constraint while $f_{ij} < 0$ means $(x_i, y_j)$ is a cannot-link constraint, with $|f_{ij}|$ denoting the confidence score of $(x_i, y_j)$ being a must-link (or cannot-link) constraint. In the following, $\mathcal{F}$ is viewed as the feasible solution set.

Similar to the representation of initial pairwise constraints on single-source data given by equation (1), the two sets of initial pairwise constraints $\mathcal{M}$ and $\mathcal{C}$ on the two-source dataset $\{\mathcal{X}, \mathcal{Y}\}$ can be denoted with a single matrix $\mathbf{Z} = \{z_{ij}\}_{N \times M}$:

$$z_{ij} = \begin{cases} +1, & (x_i, y_j) \in \mathcal{M}; \\ -1, & (x_i, y_j) \in \mathcal{C}; \\ 0, & \text{otherwise.} \end{cases} \quad (14)$$

We can find that $\mathbf{Z} \in \mathcal{F}$. Now it can be stated that the goal of the two-source constraint propagation is to find the best solution $\mathbf{F}^* \in \mathcal{F}$ based on $\mathbf{Z}$.

Although the two-source constraint propagation problem is less complicated than the original multi-source constraint propagation problem, the task of finding the best solution $\mathbf{F}^* \in \mathcal{F}$ based on $\mathbf{Z}$ is still rather difficult. Fortunately, by making vertical and horizontal observations on $\mathbf{Z}$, the two-source constraint propagation problem can be decomposed into semi-supervised learning subproblems, just as our interpretation of pairwise constraint propagation on single-source data in a semi-supervised learning perspective. Furthermore, these semi-supervised learning subproblems can be similarly merged to a single optimization problem as follows:

$$\min_{\mathbf{F}} \|\mathbf{F} - \mathbf{Z}\|_F^2 + \frac{\mu_{\mathcal{X}}}{2}\mathrm{tr}(\mathbf{F}^T\mathbf{L}_{\mathcal{X}}\mathbf{F}) + \frac{\mu_{\mathcal{Y}}}{2}\mathrm{tr}(\mathbf{F}\mathbf{L}_{\mathcal{Y}}\mathbf{F}^T), \quad (15)$$

where $\mu_{\mathcal{X}}$ (or $\mu_{\mathcal{Y}}$) denotes the regularization parameter for $\mathcal{X}$ (or $\mathcal{Y}$), and $\mathbf{L}_{\mathcal{X}}$ (or $\mathbf{L}_{\mathcal{Y}}$) denotes the normalized Laplacian matrix defined on $\mathcal{X}$ or $\mathcal{Y}$ according to equation (2). The second and third terms of the above equation denote the energy functional [18] (also known as the smoothness measure) defined over $\mathcal{X}$ and $\mathcal{Y}$, respectively. When $\mathcal{X} = \mathcal{Y}$, the above two-source constraint propagation degrades to the traditional constraint propagation on single-source data given by equation (8). In summary, we have successfully formulated both single-source and multi-source constraint propagation as minimizing a regularized energy functional.

Let $\mathcal{Q}(\mathbf{F})$ denote the objective function in equation (15). Differentiating $\mathcal{Q}(\mathbf{F})$ with respect to $\mathbf{F}$ and setting it to zero, we have the following equation:

$$\frac{\partial\mathcal{Q}}{\partial\mathbf{F}} = 2(\mathbf{F} - \mathbf{Z}) + \mu_{\mathcal{X}}\mathbf{L}_{\mathcal{X}}\mathbf{F} + \mu_{\mathcal{Y}}\mathbf{F}\mathbf{L}_{\mathcal{Y}} = 0. \quad (16)$$

Hence, the two-source constraint propagation equals to solving a Sylvester matrix equation [20]:

$$(\mathbf{I} + \mu_{\mathcal{X}}\mathbf{L}_{\mathcal{X}})\mathbf{F} + \mathbf{F}(\mathbf{I} + \mu_{\mathcal{Y}}\mathbf{L}_{\mathcal{Y}}) = 2\mathbf{Z}. \quad (17)$$

The above Sylvester matrix equation can be viewed as a generalization of the Lyapunov matrix equation (10). Since both $\mathbf{I} + \mu_{\mathcal{X}}\mathbf{L}_{\mathcal{X}}$ and $\mathbf{I} + \mu_{\mathcal{Y}}\mathbf{L}_{\mathcal{Y}}$ are positive definite, this Sylvester matrix equation has a unique solution according to [27]. A classical algorithm for the numerical solution of the Sylvester equation has been proposed in [20]. However, this algorithm incurs a large time cost. In the following, we will propose an approximate but efficient algorithm, instead of directly solving the Sylvester matrix equation.

### B. The Proposed Algorithm

Considering the strategy of solving the pairwise constraint propagation problem (8) on single-source data, we can similarly handle the two-source constraint propagation problem (15) in two optimization steps: (1) $\mathbf{F}_{\mathcal{X}}^* = \arg\min_{\mathbf{F}} \frac{1}{2}\|\mathbf{F} - \mathbf{Z}\|_F^2 + \frac{\mu_{\mathcal{X}}}{2}\mathrm{tr}(\mathbf{F}^T\mathbf{L}_{\mathcal{X}}\mathbf{F})$; (2) $\mathbf{F}^* = \arg\min_{\mathbf{F}} \frac{1}{2}\|\mathbf{F} - \mathbf{F}_{\mathcal{X}}^*\|_F^2 + \frac{\mu_{\mathcal{Y}}}{2}\mathrm{tr}(\mathbf{F}\mathbf{L}_{\mathcal{Y}}\mathbf{F}^T)$. That is, the pairwise constraint propagation is first performed on $\mathcal{X}$ and then on $\mathcal{Y}$. More notably, based on $k$-NN graphs, these two optimization problems can be solved efficiently using the label propagation technique [17].

Let $\mathbf{W}_{\mathcal{X}}$ (or $\mathbf{W}_{\mathcal{Y}}$) denote the weight matrix of the $k$-NN graph constructed on $\mathcal{X}$ (or $\mathcal{Y}$) just as Step (1) of the algorithm proposed in Section II-B. The approximate algorithm for two-source constraint propagation can be summarized as follows:

(1) Compute two matrices $\bar{\mathbf{L}}_{\mathcal{X}} = \mathbf{D}_{\mathcal{X}}^{-1/2}\mathbf{W}_{\mathcal{X}}\mathbf{D}_{\mathcal{X}}^{-1/2}$ and $\bar{\mathbf{L}}_{\mathcal{Y}} = \mathbf{D}_{\mathcal{Y}}^{-1/2}\mathbf{W}_{\mathcal{Y}}\mathbf{D}_{\mathcal{Y}}^{-1/2}$, where $\mathbf{D}_{\mathcal{X}}$ (or $\mathbf{D}_{\mathcal{Y}}$) is a diagonal matrix with its $i$-th diagonal element being the sum of the $i$-th row of $\mathbf{W}_{\mathcal{X}}$ (or $\mathbf{W}_{\mathcal{Y}}$).

(2) Iterate $\mathbf{F}_{\mathcal{X}}(t + 1) = \alpha_{\mathcal{X}}\bar{\mathbf{L}}_{\mathcal{X}}\mathbf{F}_{\mathcal{X}}(t) + (1 - \alpha_{\mathcal{X}})\mathbf{Z}$ for the pairwise constraint propagation on $\mathcal{X}$ until convergence, where $\mathbf{F}_{\mathcal{X}}(t) \in \mathcal{F}$ and $\alpha_{\mathcal{X}}$ is a parameter in the range $(0, 1)$.

(3) Iterate $\mathbf{F}_{\mathcal{Y}}(t + 1) = \alpha_{\mathcal{Y}}\mathbf{F}_{\mathcal{Y}}(t)\bar{\mathbf{L}}_{\mathcal{Y}} + (1 - \alpha_{\mathcal{Y}})\mathbf{F}_{\mathcal{X}}^*$ for the pairwise constraint propagation on $\mathcal{Y}$ until convergence, where $\mathbf{F}_{\mathcal{Y}}(t) \in \mathcal{F}$, $\mathbf{F}_{\mathcal{X}}^*$ is the limit of $\{\mathbf{F}_{\mathcal{X}}(t)\}$, and $\alpha_{\mathcal{Y}}$ is a parameter in the range $(0, 1)$.

(4) Output $\mathbf{F}^* = \mathbf{F}_{\mathcal{Y}}^*$ as the final representation of the propagated pairwise constraints, where $\mathbf{F}_{\mathcal{Y}}^*$ is the limit of $\{\mathbf{F}_{\mathcal{Y}}(t)\}$.

Similar to our analysis of the algorithm proposed in Section II-B, we can readily prove that the above algorithm for two-source constraint propagation converges to:

$$\mathbf{F}^* = (1 - \alpha_{\mathcal{X}})(1 - \alpha_{\mathcal{Y}})(\mathbf{I} - \alpha_{\mathcal{X}}\bar{\mathbf{L}}_{\mathcal{X}})^{-1}\mathbf{Z}(\mathbf{I} - \alpha_{\mathcal{Y}}\bar{\mathbf{L}}_{\mathcal{Y}})^{-1}, \quad (18)$$

where $\alpha_{\mathcal{X}} = \mu_{\mathcal{X}}/(\mu_{\mathcal{X}} + 1)$ and $\alpha_{\mathcal{Y}} = \mu_{\mathcal{Y}}/(\mu_{\mathcal{Y}} + 1)$. If we first perform pairwise constraint propagation on $\mathcal{Y}$ and then on $\mathcal{X}$, the same solution can be obtained. That is, the pairwise constraint propagation can be performed on the two data sources alternately, no matter which is first (similar to Proposition 2(ii)). Moreover, we find that the above algorithm for two-source constraint propagation has a time complexity of $O(kNM)$ which is proportional to the total number of all possible pairwise constraints. Hence, this algorithm can be considered to provide an efficient solution.

Although the above constraint propagation algorithm is limited to two-source data, it can be readily applied to multi-source data. For example, given a three-source dataset $\{\mathcal{X}, \mathcal{Y}, \mathcal{Z}\}$, this constraint propagation algorithm can be performed on three two-source datasets $\{\mathcal{X}, \mathcal{Y}\}$, $\{\mathcal{X}, \mathcal{Z}\}$, and $\{\mathcal{Y}, \mathcal{Z}\}$, respectively. The obtained three groups of results can thus be used for the retrieval with a query from a single source or a pair of queries from different sources.

### C. Application to Cross-Modal Multimedia Retrieval

When multiple sources refers to text, image, audio and so on, the output of our multi-source constraint propagation actually can be viewed as the correlation between different media sources. As we have mentioned, given the output $\mathbf{F}^* = \{f_{ij}^*\}_{N \times M}$ of two-source constraint propagation, $(x_i, y_j)$ denotes a must-link constraint if $f_{ij}^* > 0$, while $(x_i, y_j)$ denotes a cannot-link constraint if $f_{ij}^* < 0$. Considering the inherent meaning of must-link and cannot-link constraints, we can state that: $x_i$ and $y_j$ are "positively correlated" if $f_{ij}^* > 0$, while they are "negatively correlated" if $f_{ij}^* < 0$. That is, we can view $f_{ij}^*$ as the correlation coefficient between $x_i$ and $y_j$. The distinct advantage of such interpretation of $\mathbf{F}^*$ as a correlation measure is that $\mathbf{F}^*$ can thus be used for ranking on $\mathcal{Y}$ given a query $x_i$ or ranking on $\mathcal{X}$ given a query $y_j$. In fact, this is the goal of cross-modal multimedia retrieval which has drawn much attention recently [22]. That is, such challenging problem can be directly solved by our multi-source constraint propagation. It should be noted that, for cross-modal retrieval, it is not a feasible solution to combine multiple modalities as previous multi-modal retrieval methods [23]–[25] did.

In this paper, we focus on a special case of cross-modal retrieval, i.e. only text and image modalities are considered. Although various image annotation systems [28]–[30] have been developed to automatically extract semantic descriptors from images, they rely on very limited types of textual representation. Images are simply associated with keywords or class labels, without explicitly modeling of free-form text. In contrast, cross-modal retrieval is designed to deal with much more richly annotated data, motivated by the ongoing explosion of Web-based multimedia content such as news archives and Wikipedia pages (see Fig. 5). In these cases, images are related to complete text articles, and the correspondence between image and text modalities is much less direct than that provided by light annotation. At this point, cross-modal retrieval is more difficult than image annotation. To our best knowledge, little attempt has been made to directly learn the correlation between images and free-form text. One exception is the notable work of [22], where two hypotheses have been investigated for cross-modal retrieval: 1) there is a benefit to explicitly modeling the correlation between text and image modalities, and 2) this modeling is more effective with higher levels of abstraction. More concretely, the correlation between the two modalities is learned with canonical correlation analysis (CCA) [31] and abstraction is achieved by representing text and image at a more general semantic level. However, two separate steps, i.e. correlation analysis and semantic abstraction, are involved in this modeling, and the use of abstraction after CCA seems rather ad hoc.

Fortunately, this problem can be completely addressed by our multi-source constraint propagation. The semantic information (e.g. class labels) associated with images and text can be used to define the initial must-link and cannot-link constraints based on the training dataset, while the correlation between text and image modalities can be explicitly learnt by the proposed algorithm in Section III-B. That is, the correlation analysis and semantic abstraction has been successfully integrated in a unified constraint propagation framework. Our later experimental results have shown the effectiveness of such integration as compared to [22].

## IV. Experimental Results

In this section, the proposed constraint propagation algorithms are evaluated in two applications: constrained spectral clustering and cross-modal multimedia retrieval. The constraint propagation results can be directly used for cross-modal retrieval, but can only be indirectly used for constrained spectral clustering with an extra step of similarity adjustment.

### A. Constrained Spectral Clustering

We first describe the experimental setup for constrained spectral clustering, including the clustering evaluation measure and the graph construction approach. Moreover, we compare our algorithm with other closely related methods on image and UCI datasets, respectively.

*1) Experimental Setup:* For comparison, we present the results of affinity propagation (AP) [13], spectral learning (SL) [12], and semi-supervised kernel k-means (SSKK) [4], which are three constrained clustering algorithms closely related to our E$^2$CP. Here, SL and SSKK adjust only the similarities between the constrained data points, while AP and our E$^2$CP propagate the pairwise constraints throughout the entire dataset. It should be noted that AP cannot directly

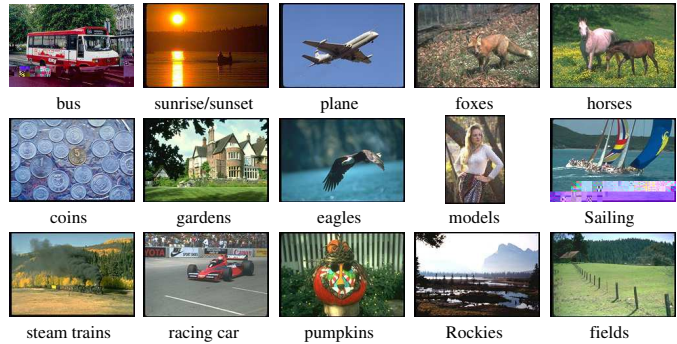

bus  sunrise/sunset  plane  foxes  horses
coins  gardens  eagles  models  Sailing
steam trains  racing car  pumpkins  Rockies  fields address multi-class problems and a time-consuming heuristic approach discussed in [13] has to be adopted. We also report the baseline results of normalized cuts (NCuts) [10], which is effectively a spectral clustering algorithm but without using pairwise constraints.

Fig. 7.  Sample images from 15 categories of the Corel dataset.

We evaluate the clustering results with the adjusted Rand (AR) index [32], [33], which has been widely used for the evaluation of clustering algorithms. The AR index measures the pairwise agreement between the obtained clustering by an algorithm and the ground truth clustering, and takes a value in the range [-1,1]. A higher AR index indicates that a higher percentage of data pairs in the obtained clustering have the same relationship (musk-link or cannot-link) as in the ground truth clustering. In the following, each experiment is randomly run 25 times, and the average AR index is obtained as the final clustering evaluation measure.

As we have mentioned, we construct a $k$-NN graph for our E$^2$CP algorithm on each dataset. To ensure a fair comparison, the same $k$-NN graph is used by all the other spectral clustering algorithms on the same dataset. For image and UCI datasets, the weight matrices of $k$-NN graphs are defined with spatial Markov kernels [19] and Gaussian kernels, respectively. That is, the spatial Markov kernels are computed on the image datasets to exploit the spatial information [19], while the Gaussian kernels are used for the UCI datasets as in [13].

*2) Results on Image Datasets:* We select two different image datasets. The first one contains 8 scene categories from MIT [34], including four man-made scenes and four natural scenes. The total number of images is 2,688. The size of each image in this Scene dataset is $256 \times 256$ pixels. The second dataset contains images from a Corel collection. We select 15 categories (see Fig. 7), and each of the categories contains 100 images. In total, this dataset has 1,500 images. The size of each image in this dataset is $256 \times 384$ pixels.

For the above two image datasets, we choose different feature sets which are introduced in [35] and [19], respectively. That is, as in [35], the SIFT descriptors are used for the Scene dataset, while, similar to [19], the joint color and Gabor features are used for the Corel dataset. These features are chosen to ensure a fair comparison with the state-of-the-art techniques. More concretely, for the Scene dataset, we extract SIFT descriptors of $16 \times 16$ pixel blocks computed over a regular grid with spacing of 8 pixels. As for the Corel dataset, we divide each image into blocks of $16 \times 16$ pixels and then extract a joint color/texture feature vector from each block. Here, the texture features are represented as the means and standard deviations of the coefficients of Gabor filters (with 3 scales and 4 orientations), and the color features are the mean values of HSV color components. Finally, for each image dataset, we perform $k$-means clustering on the extracted feature vectors to form a vocabulary of 400 visual keywords, and then define a spatial Markov kernel [19] as the weight matrix for graph construction.

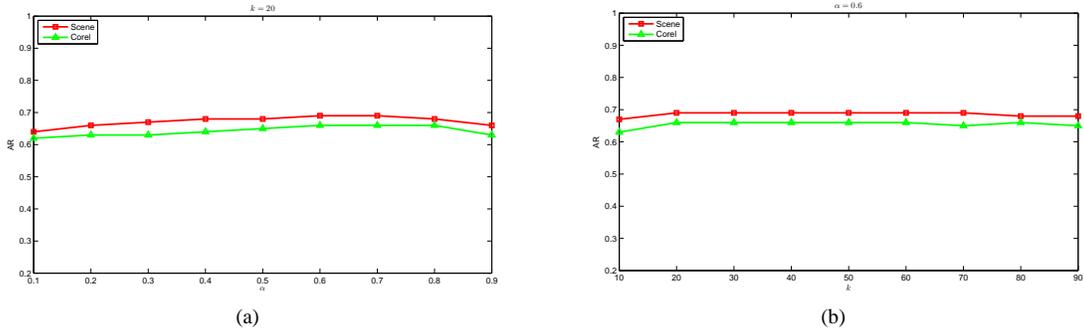

Fig. 8. Illustration of the effect of different parameters on our E²CP algorithm with 2,400 initial pairwise constraints for the two image datasets.

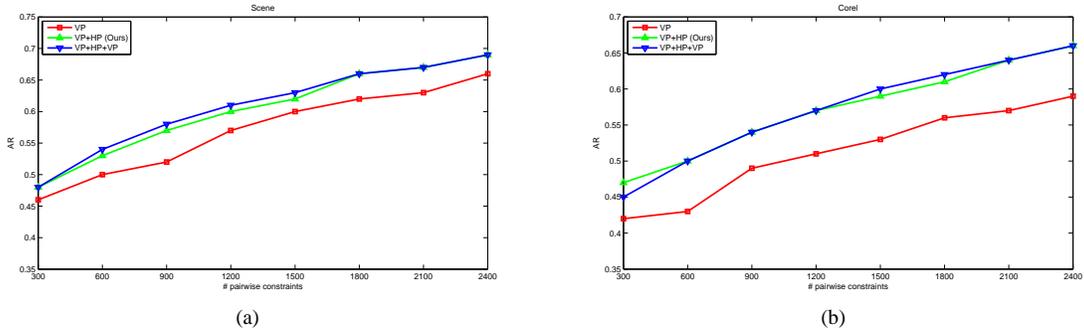

Fig. 9. Comparison between different pairwise constraint propagation approaches in single or multiple directions on the two image datasets.

In the experiments, we randomly generate the initial pairwise constraints using the ground-truth cluster labels. Moreover, we provide our E²CP algorithm with a varying number of initial pairwise constraints. In the following, we consider that the number of initial pairwise constraints ranges from 300 to 2,400. It should be noted that *the initial pairwise constraints used here are actually very sparse*. For example, the most pairwise constraints (i.e. 2,400) can be generated with only 2.6% (i.e. about 70) of the images in the Scene dataset. Here, images from the same cluster form the must-link constraints while images from different clusters form the cannot-link constraints. When such few labeled images are initially provided, it is not feasible to select the parameters by cross-validation for our E²CP algorithm. Hence, we set $\alpha = 0.6$ and $k = 20$ empirically (similar to the parameter selection for many semi-supervised learning algorithms). In fact, as shown in Fig. 8, our E²CP algorithm is not sensitive to these two parameters, and we select a relatively smaller value for $k$ to ensure its efficient running.

Considering that our E²CP algorithm has two key steps: vertical constraint propagation and horizontal constraint propagation, we need to demonstrate the importance of the horizontal constraint propagation as a supplement to the verti-cal constraint propagation. Hence, we compare the following three constraint propagation approaches: only vertical propagation (VP), both vertical and horizontal propagation (VP+HP), and VP+HP followed with vertical propagation again (VP+HP+VP). The results are shown in Fig. 9. The immediate observation is that the horizontal propagation is crucial for our E²CP algorithm (see VP vs. VP+HP). Moreover, we also find that the vertical propagation does not need to be performed more than one time, since extra propagation leads to very minor improvements (see VP+HP vs. VP+HP+VP).

It should be noted that the propagated pairwise constraints by our E²CP algorithm can not be directly used for spectral clustering and they need to be first exploited for similarity adjustment according to equation (13). The effectiveness of this similarity adjustment approach has been preliminarily verified by Proposition 3. The further verification is shown in Fig. 10, where NewWeight1 means $\tilde{w}_{ij} = 1 + f_{ij}^*$, NewWeight2 means $\tilde{w}_{ij} = (1 + f_{ij}^*)w_{ij}$, and NewWeight3 means equation (13). Here, NewWeight1 defines the new weight matrix only based on the output of our E²CP algorithm (i.e. the similarity adjustment step is actually ignored), while both NewWeight2 and NewWeight3 additionally consider the original weight matrix. The significant gains achieved by NewWeight3 over NewWeight1 show that the similarity adjustment step according to equation (13) is crucial for the success of the overall method. Moreover, as compared with NewWeight2, our approach is shown to perform much better, which also provides further support for Proposition 3.

Finally, we compare our E²CP algorithm with other closely related methods on the two image datasets. To verify that the approximate solution obtained by our E²CP is comparable to that of the Lyapunov equation (10), we also make comparison with the constraint propagation approach by directly solving the Lyapunov equation with $\mu = \alpha/(1 - \alpha)$ and $\mathbf{L} = \mathbf{I} - \bar{\mathbf{L}}$ (denoted as LYAP). The overall clustering results are shown in Fig. 11, and the running time taken on the Scene dataset is also listed in Table I. Here, we run all the clustering algorithms (Matlab code) on a PC with 4GB RAM and two 2.66 GHz CPUs. The immediate observation is that the performance of our $E^2CP$ algorithm is comparable (even slightly better) to that of LYAP. Considering that our $E^2CP$ algorithm incurs much less time cost, we prefer it to LYAP in practice.

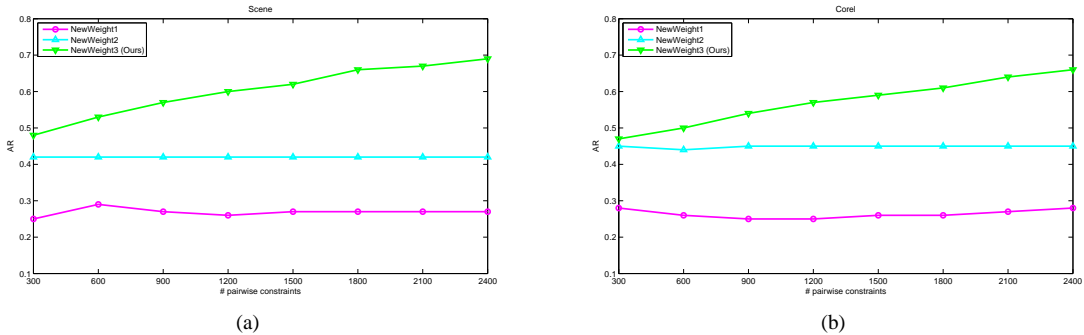

Fig. 10. Comparison between different approaches to computing the new weight matrix for constrained spectral clustering on the two image datasets.

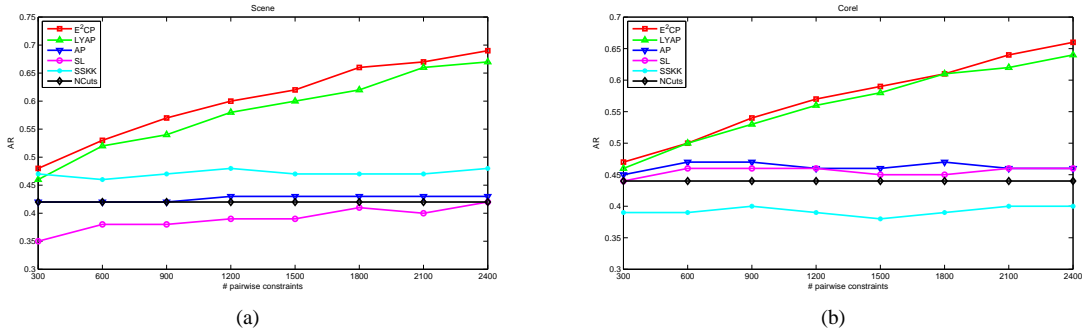

Fig. 11. The clustering results on the two image datasets by different algorithms when a varying number of pairwise constraints are initially provided.

TABLE I
THE RUNNING TIME TAKEN BY DIFFERENT ALGORITHMS ON THE SCENE DATASET WHEN 2400 PAIRWISE CONSTRAINTS ARE INITIALLY PROVIDED.

| Algorithms | $E^2CP$ | LYAP | AP | SL | SSKK | NCuts |
|---|---|---|---|---|---|---|
| Time (sec.) | 12 | 272 | 21 | 9 | 35 | 6 |

TABLE II
FOUR UCI DATASETS USED IN THE EXPERIMENTS. THE FEATURES ARE FIRST NORMALIZED TO THE RANGE [-1, 1] FOR ALL THE DATASETS.

| Datasets | Wine | Ionosphere | Soybean | WDBC |
|---|---|---|---|---|
| # samples | 178 | 351 | 47 | 569 |
| # features | 13 | 34 | 35 | 30 |
| # clusters | 3 | 2 | 4 | 2 |

The further observation on Fig. 11 shows that our $E^2CP$ performs consistently better than other closely related methods. That is, the effectiveness of our exhaustive constraint propagation approach to exploiting pairwise constraints for spectral clustering has been verified by the promising performance of our $E^2CP$. In contrast, SL and SSKK perform unsatisfactorily, and, in some cases, their performance has been degraded to that of NCuts. This may be due to that by merely adjusting the similarities only between the constrained images, these approaches have not fully utilized the additional supervisory or prior information inherent in the constrained images, and hence can not discover the complex structures hidden in the challenging image datasets. Although AP can also propagate pairwise constraints throughout the entire dataset like our $E^2CP$, the heuristic approach discussed in [13] may not ad-dress multi-class problems for the challenging image datasets, which thus leads to unsatisfactory results. Moreover, another important observation is that the performance improvement by our $E^2CP$ with respect to NCuts becomes more obvious when more pairwise constraints are provided, while this is not the case for AP, SL or SSKK. In other words, the pairwise constraints has been exploited more exhaustively and effectively by our pairwise constraint propagation.

Besides the above advantages over other closely related methods, our $E^2CP$ has another advantage in terms of time cost. That is, as shown in Table I, the running time of our $E^2CP$ is comparable to that of the constrained spectral clustering algorithm without using constraint propagation (i.e. SL). Moreover, as for the two constraint propagation approaches (i.e. $E^2CP$ and AP), our $E^2CP$ runs faster than AP, particularly for multi-class problems.

*3) Results on UCI Datasets:* We further conduct experiments on four UCI datasets, which are described in Table II. The UCI datasets have been widely used to evaluate clustering and classification algorithms in machine learning. Here, as in [13], the Gaussian kernel is defined on each UCI dataset for computing the weight matrix during graph construction. The experimental setup and parameter selection on the UCI datasets are similar to those for the image datasets. Since we have verified the effectiveness of each important component of our E²CP algorithm on the image datasets, we only make comparison with other closely related methods on the UCI datasets. The clustering results are shown in Fig. 12.

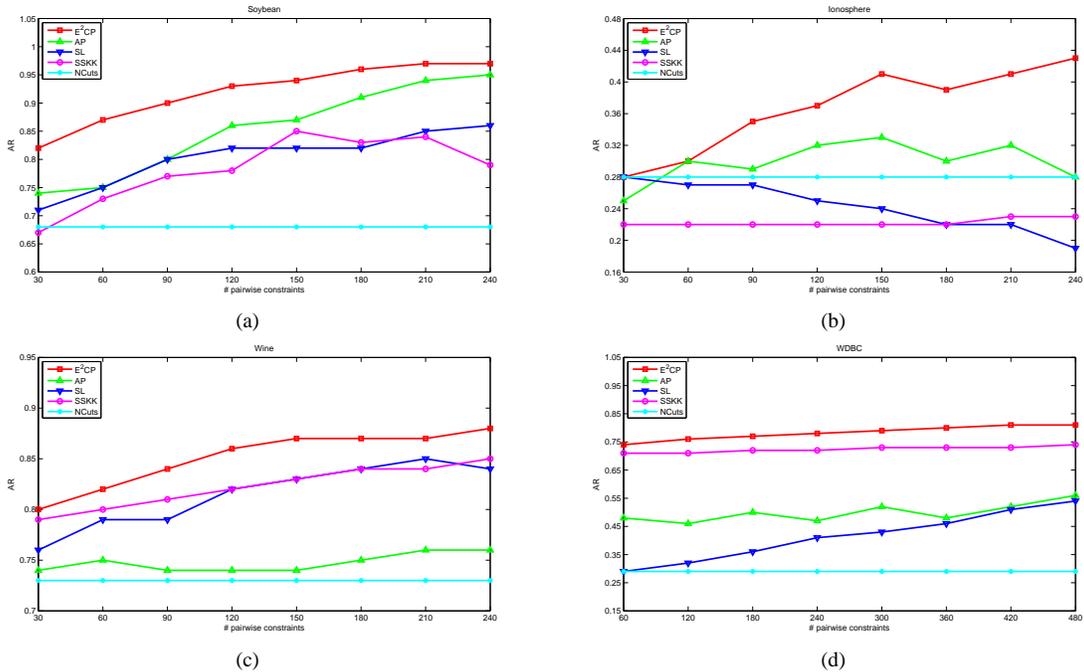

Fig. 12. The clustering results on the four UCI datasets by different algorithms when a varying number of pairwise constraints are initially provided.

Again, we can find that our constraint propagation approach (i.e. E²CP) achieves improved performance in most cases. Moreover, the other three constrained clustering approaches (i.e. AP, SL, and SSKK) are shown to have generally benefited from the pairwise constraints as compared to NCuts. This observation is different from that on the image datasets. As we have mentioned, this may be due to that, considering the complexity of the image datasets, a more exhaustive propagation (like our E²CP) of the pairwise constraints is needed in order to fully utilize the inherent supervisory information provided by the pairwise constraints. Our experimental results have also demonstrated that an exhaustive propagation of the pairwise constraints in the UCI datasets through our E²CP leads to improved clustering performance over the other three constrained clustering approaches (i.e. AP, SL, and SSKK).

### B. Cross-Modal Multimedia Retrieval

In this subsection, our multi-source constraint propagation (MSCP) algorithm is evaluated in the challenging application of cross-modal multimedia retrieval. We focus on comparing our MSCP algorithm with the state-of-the-art approach [22], since they both consider not only correlation analysis but also semantic abstraction for text and image modalities. To verify that the approximate solution obtained by our MSCP is comparable to that of the Sylvester equation (17), we also make comparison with the approach by directly solving the Sylvester equation with $\mu_{\mathcal{X}} = \alpha_{\mathcal{X}}/(1 - \alpha_{\mathcal{X}})$, $\mu_{\mathcal{Y}} = \alpha_{\mathcal{Y}}/(1 - \alpha_{\mathcal{Y}})$, $\mathbf{L}_{\mathcal{X}} = \mathbf{I} - \bar{\mathbf{L}}_{\mathcal{X}}$, and $\mathbf{L}_{\mathcal{Y}} = \mathbf{I} - \bar{\mathbf{L}}_{\mathcal{Y}}$ (denoted as SYLV).

*1) Experimental Setup:* We conduct experiments on a cross-modal retrieval benchmark dataset [22], which contains a total of 2,866 documents. Each document is actually a text-image pair, annotated with a label from the vocabulary of 10 semantic classes. Here, it should be noted that these documents are derived from Wikipedia's "featured articles". The original featured articles are categorized by Wikipedia's editors into 29 categories, and these category labels are assigned to both the text and image components of each article. Since some of the categories are very scarce, we only consider the 10 most populated ones, just as [22].

In this dataset, the text representation for each document is derived from a latent Dirichlet allocation (LDA) model with 10 latent topics, while the image representation is based on a bag-of-words model with 128 visual words learnt from the extracted SIFT descriptors, which is exactly the same as [22]. Moreover, following the strategy of [22], the normalized correlation measure is used to define the similarity matrix for both text and image representation.

The benchmark dataset [22] is split into a training set of 2,173 documents and a test set of 693 documents. The initial pairwise constraints for our MSCP algorithm are derived from the class labels of the training documents. The performance of our MSCP algorithm is evaluated on the test set. Here, two tasks are considered: text retrieval using an image query, and image retrieval using a text query. In the following, these two tasks are denoted as "Image Query" and "Text Query", respectively. For each task, the retrieval results are measured with mean average precision (MAP) which has been widely used in the image retrieval literature [36].

Let $\mathcal{X}$ denote the text source and $\mathcal{Y}$ denote the image source. For our MSCP algorithm, we construct $k$-NN graphs on $\mathcal{X}$ and $\mathcal{Y}$ with the same $k$. As we have mentioned, in the application of constrained spectral clustering on image datasets, at most 70 labeled images are initially provided for our E²CP algorithm and thus we can not select the parameters by cross-validation on such a small training set. However, for the cross-modal retrieval benchmark dataset [22], we have a much larger training set (of the size 2,173) and all the parameters (i.e. $\alpha_\mathcal{X}$, $\alpha_\mathcal{Y}$, and $k$) can be selected by fivefold cross-validation for our MSCP algorithm. More concretely, according to Fig. 13, we set the three parameters as: $\alpha_\mathcal{X} = 0.1$, $\alpha_\mathcal{Y} = 0.025$, and $k = 90$. We can also find that our MSCP algorithm is not sensitive to these parameters.

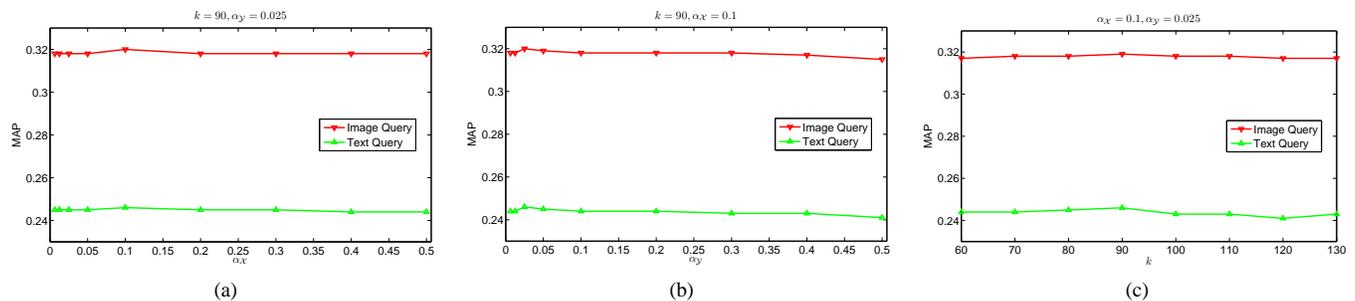

Fig. 13. The cross-modal retrieval results by fivefold cross-validation on the training set for our MSCP algorithm.

TABLE III
THE CROSS-MODAL RETRIEVAL RESULTS ON THE TEST SET MEASURED BY
MAP SCORES.

| Methods | CA [22] | SA [22] | CA+SA [22] | SYLV | MSCP |
|---|---|---|---|---|---|
| Image Query | 0.249 | 0.225 | 0.277 | **0.329** | **0.329** |
| Text Query | 0.196 | 0.223 | 0.226 | **0.256** | **0.256** |

*2) Cross-Modal Retrieval Results:* As reported in [22], both correlation analysis (CA) and semantic abstraction (SA) play an important role in cross-modal retrieval (also see Table III). However, these two steps are completely separate in this modeling, and the use of semantic abstraction after correlation analysis (i.e. CA+SA) seems rather ad hoc.

In fact, as we have mentioned in Section III-C, these two steps can be seamlessly integrated in a unified multi-source constraint propagation framework. That is, the semantic information (e.g. class labels) associated with images and text can be used to define the initial must-link and cannot-link constraints based on the training set, while the correlation between text and image modalities can be explicitly learnt by our MSCP algorithm proposed in Section III-B. The cross-modal retrieval results listed in Table III have shown the effectiveness of such integration as compared to the state-of-the-art approach (CA+SA) [22]. This means that the initial supervisory information provided for cross-modal retrieval can be more exhaustively utilized by our MSCP algorithm through pairwise constraint propagation across text and image modalities, which is similar to the observations in the application of constrained spectral clustering.

Moreover, we can also find that the performance of our MSCP algorithm is as good as that of SYLV. Considering that our MSCP algorithm incurs much less cost (34 seconds vs. 385 seconds), we prefer it to SYLV in practice. In fact, this conclusion is the same as that made for our E²CP algorithm as compared to LYAP on single-source data.

## V. CONCLUSIONS

We have investigated the pairwise constraint propagation problem in a semi-supervised learning perspective. By decomposing this challenging problem into a set of independent semi-supervised learning subproblems, we have successfully formulated it as minimizing a regularized energy functional. More importantly, these semi-supervised learning subproblems can be solved efficiently and in parallel using the label propagation technique based on k-nearest neighbor graphs. The resulting exhaustive set of propagated pairwise constraints are exploited for similarity adjustment in the application of constrained spectral clustering. It is worth noting that we have first clearly shown how pairwise constraints are propagated throughout the entire dataset. The proposed approach on single-source data is further extended to more challenging constraint propagation on multi-source data with an important application to cross-modal multimedia retrieval. Extensive results have shown that our exhaustive and efficient constraint propagation approach can achieve superior performance in both constrained spectral clustering and cross-modal retrieval. For future work, our approach will also be used to improve the performance of other graph-based methods by exhaustively exploiting the pairwise constraints.

## ACKNOWLEDGEMENTS

This work was supported in part by the National Natural Science Foundation of China under Grant Nos. 60873154 and 61073084, by the City University of Hong Kong under Grant No. 7008040, and by CityU matching grant to Australian Linkage grant No. 9678017.

## REFERENCES

[1] K. Wagstaff, C. Cardie, S. Rogers, and S. Schroedl, "Constrained k-means clustering with background knowledge," in *Proc. ICML*, 2001, pp. 577–584.

[2] D. Klein, S. Kamvar, and C. Manning, "From instance-level constraints to space-level constraints: Making the most of prior knowledge in data clustering," in *Proc. ICML*, 2002, pp. 307–314.

[3] S. Basu, M. Bilenko, and R. Mooney, "A probabilistic framework for semi-supervised clustering," in *Proc. SIGKDD*, 2004, pp. 59–68.

[4] B. Kulis, S. Basu, I. Dhillon, and R. Mooney, "Semi-supervised graph clustering: A kernel approach," in *Proc. ICML*, 2005, pp. 457–464.

[5] E. Xing, A. Ng, M. Jordan, and S. Russell, "Distance metric learning with application to clustering with side-information," in *Advances in Neural Information Processing Systems 15*, 2003, pp. 505–512.

[6] S. Hoi, W. Liu, M. Lyu, and W.-Y. Ma, "Learning distance metrics with contextual constraints for image retrieval," in *Proc. CVPR*, 2006, pp. 2072–2078.

[7] W. Liu, S. Ma, D. Tao, J. Liu, and P. Liu, "Semi-supervised sparse metric learning using alternating linearization optimization," in *Proc. KDD*, 2010, pp. 1139–1148.

[8] A. Ng, M. Jordan, and Y. Weiss, "On spectral clustering: Analysis and an algorithm," in *Advances in Neural Information Processing Systems 14*, 2002, pp. 849–856.

[9] U. von Luxburg, "A tutorial on spectral clustering," *Statistics and Computing*, vol. 17, no. 4, pp. 395–416, 2007.

[10] J. Shi and J. Malik, "Normalized cuts and image segmentation," *IEEE Trans. Pattern Analysis and Machine Intelligence*, vol. 22, no. 8, pp. 888–905, 2000.

[11] O. Veksler, "Star shape prior for graph-cut image segmentation," in *Proc. ECCV*, 2008, pp. 454–467.

[12] S. Kamvar, D. Klein, and C. Manning, "Spectral learning," in *Proc. IJCAI*, 2003, pp. 561–566.

[13] Z. Lu and M. Carreira-Perpinan, "Constrained spectral clustering through affinity propagation," in *Proc. CVPR*, 2008.

[14] Z. Li, J. Liu, and X. Tang, "Pairwise constraint propagation by semidefinite programming for semi-supervised classification," in *Proc. ICML*, 2008, pp. 576–583.

[15] S. Yu and J. Shi, "Segmentation given partial grouping constraints," *IEEE Trans. Pattern Analysis and Machine Intelligence*, vol. 26, no. 2, pp. 173–183, 2004.

[16] Z. Lu and H. Ip, "Constrained spectral clustering via exhaustive and efficient constraint propagation," in *Proc. ECCV*, vol. 6, 2010, pp. 1–14 (Oral Presentation).

[17] D. Zhou, O. Bousquet, T. Lal, J. Weston, and B. Schölkopf, "Learning with local and global consistency," in *Advances in Neural Information Processing Systems 16*, 2004, pp. 321–328.

[18] X. Zhu, Z. Ghahramani, and J. Lafferty, "Semi-supervised learning using Gaussian fields and harmonic functions," in *Proc. ICML*, 2003, pp. 912–919.

[19] Z. Lu and H. Ip, "Image categorization by learning with context and consistency," in *Proc. CVPR*, 2009, pp. 2719–2726.

[20] R. Bartels and G. Stewart, "Solution of the matrix equation $AX + XB = C$," *Communications of the ACM*, vol. 15, no. 9, pp. 820–826, 1972.

[21] Z. Gajic and M. Qureshi, Eds., *Lyapunov Matrix Equation in System Stability and Control*. Academic Press, 1995.

[22] N. Rasiwasia, J. Costa Pereira, E. Coviello, G. Doyle, G. Lanckriet, R. Levy, and N. Vasconcelos, "A new approach to cross-modal multimedia retrieval," in *Proc. ACM Multimedia*, 2010, pp. 251–260 (Best Student Paper).

[23] E. Bruno, N. Moenne-Loccoz, and S. Marchand-Maillet, "Design of multimodal dissimilarity spaces for retrieval of video documents," *IEEE Trans. Pattern Analysis and Machine Intelligence*, vol. 30, no. 9, pp. 1520–1533, 2008.

[24] R. Bekkerman and J. Jeon, "Multi-modal clustering for multimedia collections," in *Proc. CVPR*, 2007, pp. 1–8.

[25] C. Snoek and M. Worring, "Multimodal video indexing: A review of the state-of-the-art," *Multimedia Tools and Applications*, vol. 25, no. 1, pp. 5–35, 2005.

[26] M. Law, A. Topchy, and A. Jain, "Clustering with soft and group constraints," in *Proceedings of the Joint IAPR International Workshop on Structural, Syntactic, and Statistical Pattern Recognition*, 2004, pp. 662–670.

[27] P. Lancaster, "Explicit solutions of linear matrix equations," *SIAM Review*, vol. 12, no. 4, pp. 544–566, 1970.

[28] J. Li and J. Wang, "Automatic linguistic indexing of pictures by a statistical modeling approach," *IEEE Trans. Pattern Analysis and Machine Intelligence*, vol. 25, no. 9, pp. 1075–1088, Sept. 2003.

[29] S. Feng, R. Manmatha, and V. Lavrenko, "Multiple Bernoulli relevance models for image and video annotation," in *Proc. CVPR*, vol. 2, 2004, pp. 1002–1009.

[30] Z. Lu, H. Ip, and Y. Peng, "Contextual kernel and spectral methods for learning the semantics of images," *IEEE Trans. Image Processing*, vol. 20, no. 6, pp. 1739–1750, 2011.

[31] H. Hotelling, "Relations between two sets of variates," *Biometrika*, vol. 28, no. 3-4, pp. 321–377, 1936.

[32] L. Hubert and P. Arabie, "Comparing partitions," *Journal of Classification*, vol. 2, no. 1, pp. 193–218, 1985.

[33] Z. Lu, Y. Peng, and J. Xiao, "From comparing clusterings to combining clusterings," in *Proc. AAAI*, 2008, pp. 665–670.

[34] A. Oliva and A. Torralba, "Modeling the shape of the scene: A holistic representation of the spatial envelope," *International Journal of Computer Vision*, vol. 42, no. 3, pp. 145–175, 2001.

[35] A. Bosch, A. Zisserman, and X. Muñoz, "Scene classification via pLSA," in *Proc. ECCV*, 2006, pp. 517–530.

[36] N. Rasiwasia, P. Moreno, and N. Vasconcelos, "Bridging the gap: Query by semantic example," *IEEE Trans. Multimedia*, vol. 9, no. 5, pp. 923–938, 2007.

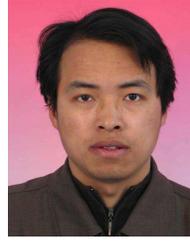

**Zhiwu Lu** received the M.Sc. degree in applied mathematics from Peking University, Beijing, China in 2005, and the Ph.D. degree in computer science from City University of Hong Kong in 2011.

From July 2005 to August 2007, he was a software engineer with Founder Corporation, Beijing, China. From September 2007 to June 2008, he was a research assistant with the Institute of Computer Science and Technology, Peking University. Since March 2011, he has become an assistant professor with the Institute of Computer Science and Technology, Peking University. He has published over 30 papers in international journals and conference proceedings including TIP, TSMC-B, TMM, AAAI, ICCV, CVPR, ECCV, and ACM-MM. His research interests lie in pattern recognition, machine learning, multimedia information retrieval, and computer vision.

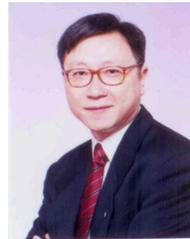

**Horace H.S. Ip** received the B.Sc. (first-class honors) degree in applied physics and the Ph.D. degree in image processing from the University College London, London, U.K., in 1980 and 1983, respectively.

Currently, he is the Chair Professor of computer science, the Founding Director of the Centre for Innovative Applications of Internet and Multimedia Technologies (AIMtech Centre), and the Acting Vice-President of City University of Hong Kong, Kowloon, Hong Kong. He has published over 200 papers in international journals and conference proceedings. His research interests include pattern recognition, multimedia content analysis and retrieval, virtual reality, and technologies for education. He is a Fellow of the Hong Kong Institution of Engineers, a Fellow of the U.K. Institution of Electrical Engineers, and a Fellow of the IAPR.

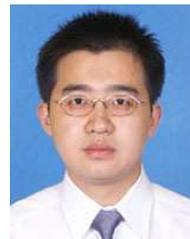

**Yuxin Peng** received the Ph.D. degree in computer science and technology from Peking University, Beijing, China, in 2003.

He joined the Institute of Computer Science and Technology, Peking University, as an assistant professor in 2003 and was promoted to a professor in 2010. From 2003 to 2004, he was a visiting scholar with the Department of Computer Science, City University of Hong Kong. His current research interests include content-based video retrieval, image processing, and pattern recognition.